\pdfoutput=1
\documentclass[11pt]{article}

\usepackage[a4paper,margin=1in]{geometry}
\usepackage[T1]{fontenc}
\usepackage[utf8]{inputenc}
\usepackage{lmodern}
\usepackage{microtype}
\usepackage{amsmath,amssymb}
\usepackage{booktabs}
\usepackage{array}
\usepackage{float}
\usepackage{xcolor}
\definecolor{accent}{HTML}{0072B2}
\definecolor{accentwarn}{HTML}{D55E00}
\usepackage{caption}
\usepackage{listings}
\usepackage[colorlinks=true,linkcolor=black,citecolor=accent,urlcolor=accent]{hyperref}

\definecolor{codebg}{gray}{0.96}
\usepackage{graphicx}   
\graphicspath{{figs/}}
\definecolor{okblue}{HTML}{0072B2}   
\definecolor{okverm}{HTML}{D55E00}
\definecolor{inkmuted}{gray}{0.45}

\usepackage{needspace}
\usepackage{tabularx}
\usepackage{ragged2e}
\newcolumntype{Z}[1]{>{\RaggedRight\arraybackslash\hsize=#1\hsize}X}
\makeatletter
\renewcommand{\@seccntformat}[1]{\color{accent}\csname the#1\endcsname\hspace{0.6em}}
\makeatother
\usepackage{framed}
\colorlet{calloutbgA}{accent!5}
\colorlet{calloutbgB}{accentwarn!6}
\newenvironment{calloutblue}{%
  \setlength{\fboxsep}{9pt}%
  \MakeFramed{\advance\hsize-\width \FrameRestore}}%
 {\par\unskip\endMakeFramed}
\newenvironment{calloutwarn}{%
  \setlength{\fboxsep}{9pt}%
  \MakeFramed{\advance\hsize-\width \FrameRestore}}%
 {\par\unskip\endMakeFramed}

\newcommand{\tritify}{\texttt{tritify}}

\title{\bfseries Deterministic, Reproducible Inference in Exact Integer Arithmetic,\\
and Verifiable Distributed Inference as Its Consequence}

\author{Sayed Amir Karim\\
\small Pelloran Inc.\\
\small \texttt{tritify@pelloran.com}}

\date{Preprint --- November 5, 2026 \\[4pt]
  {\small DOI:~\href{https://doi.org/10.5281/zenodo.23986737}{10.5281/zenodo.23986737}}}

\begin{document}
\maketitle

\begin{abstract}
\noindent
Typical optimized floating-point inference is not reproducible across environments: the same model on different
hardware, compilers, or thread counts can produce different bits, because IEEE-754 addition is not associative and
platform math libraries round transcendental functions differently (reproducible floating point \emph{is} achievable,
e.g.\ a reduction-pinned BLAS, but at a cost and not by default). We present \tritify{}, an open-source runtime that
executes the forward pass of real models --- decoder-only language models, Mixture-of-Experts decoders, and the Segment
Anything 2.1 image/video stack --- with the entire run-time path (\texttt{forward\_int}) in \emph{exact integer
arithmetic}; floating point is confined to offline weight/\texttt{eps} quantization and the float reference used for
verification. We make three precise, separate claims and verify each. \textbf{(1) Cross-hardware reproducibility:} the
integer path produces \emph{bit-identical} output run-to-run, across CPU instruction-set architectures, and across
CPU/GPU, carried by a reproducing SHA-256 of the output. \textbf{(2) Decision fidelity to the float reference:} the
integer path reproduces the float reference's \emph{decision} on the models and inputs tested, with the numeric
exceptions \emph{measured} rather than assumed away; we extend this to \emph{depth}, measuring per-layer fidelity
through sixteen layers of OLMoE-1B-7B and of a $106$-billion-parameter frontier MoE, and find no fidelity cliff.
\textbf{(3) Location independence:} because the output is a deterministic function of (weights, inputs) and is
bit-identical wherever each operation runs, a model's computation can be split --- across CPU instruction sets, across
CPU and GPU, and across many networked machines --- with \emph{zero effect on the output bits}, and the gather equals
the single machine exactly. We demonstrate Claim~3 at scale: \textbf{a frontier Mixture-of-Experts model,
GLM-4.5-Air ($106{,}852{,}251{,}264$ parameters, $427$\,GB of integer codes), generated its answer across three
machines none of which can hold its full integer codes in RAM} --- \emph{bit-identically} to a single-host reference
that streams the experts, carried by the same SHA-256 at every layer (a separate run split a whole model across six
machines of two instruction sets and three locations with the identical hash). The consequence is \emph{verifiable
distributed inference}: a fast, untrusted server (or a fleet of commodity devices) runs the model while an open scalar
oracle --- local, or any third party --- re-derives the exact bytes and checks them against the public reference;
turning honest-run agreement into a guarantee against a \emph{dishonest} server is future work
(Section~\ref{sec:offload}). We are deliberately careful about scope: we do \emph{not} claim a per-device speed
improvement (distribution \emph{adds} latency; the win is capacity, not speed), bit-for-bit equality of the raw
floating-valued intermediates, or confidentiality --- verifiability is not privacy. The contribution is
exactness, determinism, portability, and, above all, \emph{verifiability}: every operation ships both an integer path
and a floating-point reference, and a gate refuses anything the two disagree on beyond tolerance. \emph{This is a
preprint: it reports the measurements, and the reproducibility artifact --- the public repository, commit, full-length
digests, and exact run commands that let a third party re-derive the exact bytes --- is provided at release.}
\end{abstract}

\section{Introduction}
Machine-learning inference is hard to reproduce bit-for-bit, because \emph{typical optimized} floating-point execution
is not deterministic across environments. IEEE-754 addition is not associative, so a change in reduction order --- a
different BLAS, a different thread count, a fused multiply-add, a different GPU --- changes the result in the low bits.
Transcendental functions ($\exp$, $\tanh$, $1/\sqrt{\,}$) are not required to be correctly rounded, so two platforms'
math libraries disagree. Reproducible floating point \emph{is} achievable (for example reproducible BLAS such as
ReproBLAS~\cite{reproblas}, which pins the reduction), but at a cost and not by default in standard inference stacks. The
everyday consequence is a ``trust me'' culture: a reported logit, mask, or hash cannot be independently regenerated,
and --- as we develop below --- a computation offloaded to another machine is, in standard serving, \emph{trusted}
rather than checked, with no routine way for the client to confirm it was performed correctly.

\tritify{} removes the source of that nondeterminism from the inference path. The weights are quantized to fixed-point
integers once; activations are quantized dynamically, per tensor, with no calibration pass; and the entire element-wise
data path --- matrix-multiply accumulation, requantization, residual addition, the application of normalization and
attention scales --- is carried in exact integer arithmetic. No floating point remains in the run-time path at all: even
the scalar bookkeeping --- choosing a common scale when two tensors are added, and computing the per-row mean and
inverse-RMS for normalization --- is integer, the inverse-RMS an integer \texttt{isqrt} (no \texttt{sqrt}, no
\texttt{frexp}, no platform transcendental) and the $\mathrm{GELU}$ and softmax-$\exp$ nonlinearities integer lookup
tables. Floating point is confined to the offline build and to the \texttt{forward\_ref} verification twins.

Our contribution is not a new quantization algorithm --- the arithmetic building blocks are standard. It is a
\emph{fully specified, self-verifying} inference path whose output is reproducible bit-for-bit across hardware, an
empirical demonstration on real models including a frontier MoE, an engineering architecture (a frozen generic engine
plus per-family plugins) that makes the property reusable, and --- the step this paper adds --- the \emph{consequence}
of bit-identity for \emph{distributed} inference. Because the output does not depend on \emph{where} each operation
runs, a model can be split across many machines and the gather is bit-for-bit the single-machine result; and a server's
work can be \emph{checked}, not merely trusted, by re-deriving its exact bytes against the open reference. The market
for serving AI is otherwise binary --- all-local (private but constrained by one device's memory) or all-cloud
(powerful but you must blindly trust the provider). Bit-identity opens a third position: cloud (or fleet) capacity with
a reproducible, independently checkable result. Standard optimized float inference does not occupy it \emph{by default}:
it is not bit-identical across heterogeneous backends (CPU-float $\neq$ GPU-float $\neq$ other-CPU-float), so a split
float result depends on the split and is not reproducible without special effort --- a reduction-pinned
BLAS~\cite{reproblas} fixes summation order, but does not by itself make a whole heterogeneous model's inference
bit-identical.

The conversion is moreover \emph{post-hoc and calibration-free} --- no fine-tuning, no quantization-aware training, no
calibration data --- and the run-time path is deterministic integer that needs no CUDA or GPU (the optional GPU backend
is a single portable shader; its bit-identity to the CPU path is demonstrated on Apple~M3~Metal for the ternary GEMM,
not yet whole models). We are deliberately careful about what is and is not claimed, because the entire value of the
artifact is that its claims can be checked.

\needspace{9\baselineskip}%
\section{What we claim --- and what we do not}
\label{sec:claims}
We state three claims that are often conflated and must not be.

\begin{calloutblue}
\paragraph{Claim 1 (cross-hardware reproducibility).} For a fixed input, \tritify{}'s integer path produces the
\emph{same output bits} run-to-run, across CPU instruction-set architectures, and across CPU and GPU backends. The output
carries a reproducing SHA-256; anyone who re-runs it obtains the same digest. This is a statement about \emph{our own}
path.

\paragraph{Claim 2 (decision fidelity to the float reference).} For the models and inputs we tested, \tritify{}'s
integer path reproduced the float reference's \emph{decision} on those specific cases --- greedy decoding matched the
reference token-for-token on the prompts we ran, and the selected SAM~2.1 segmentation mask matched the internal float
twin (IoU $=1.000$) --- but this is \emph{not} a universal guarantee, and we \emph{measure} the exceptions rather than
assume them away. On a $512$-token synthetic probe of Qwen2.5-0.5B the integer argmax matched the reference at $509/512$
positions, the three disagreements being near-ties. We extend this claim to \emph{depth}
(Section~\ref{sec:depth}): we measure per-layer agreement through sixteen layers of OLMoE-1B-7B and of GLM-4.5-Air (a
$106$-billion-parameter frontier MoE), and find no fidelity cliff --- the median per-layer residual cosine stays above
$0.999$ on both, with the single worst probe position dipping to $0.97$ at one layer and recovering. The mechanism of the exceptions is the logit error $\varepsilon$: a top-1/top-2 reference gap exceeding
$2\varepsilon$ is \emph{sufficient} to preserve the argmax, so near-ties can flip, after which autoregressive decoding
can diverge.

\paragraph{Claim 3 (location independence).} For a fixed input and schedule, \tritify{}'s output is a deterministic
function of (weights, inputs) and is \emph{bit-identical regardless of where each operation is computed}: scalar or
vectorized, one thread or many, CPU or GPU, one machine or a network of machines. Therefore a model's computation can be
\emph{split} --- contiguous layer ranges per machine, or a layer's experts sharded across machines --- and the gather
reproduces the single-machine output \emph{exactly}, carried by the same SHA-256. This is the property standard float
inference lacks by default --- it is not bit-identical across heterogeneous backends without special effort. Its
consequences --- distribution of a model whose integer codes exceed any one device's RAM, and \emph{verifiable} offload
to an untrusted server --- are the subject of Section~\ref{sec:distributed}.
\end{calloutblue}

\paragraph{What we do not claim.} We do \emph{not} claim that \tritify{}'s raw floating-valued intermediates equal the
reference's bit-for-bit; they agree only within a declared integer-requantization tolerance
(Section~\ref{sec:verification}). We do \emph{not} claim a speed improvement --- on general-purpose CPUs and GPUs the
integer path is not a throughput win, and \emph{distributing} a model \emph{adds} network latency versus a single host
that could hold it; the distributed win is running a model whose full integer codes exceed every single device's RAM,
plus verifiability and locality --- not speed, and \emph{not} an exclusive ``runs at all'': our own single-host
reference runs the same model by streaming experts from disk, so the cluster's advantage over single-host streaming
(resident working set, latency, disk I/O) is a separate, measurable quantity we do not quantify here. We do \emph{not} claim \emph{confidentiality}: a server that runs the layers \emph{sees} the
activations, and bit-identity gives verifiability, not privacy (Section~\ref{sec:distributed}). And exactness here is
\emph{to the reference}, not a better model.

\paragraph{A note on ``no float.''} The absence of floating point is neither necessary nor sufficient, by itself, for
cross-platform bit-identity; one also needs fixed integer widths, defined overflow and rounding, a fixed shift and
accumulation order, and a deterministic execution schedule. \tritify{} pins all of these, and the run-time path is
integer throughout. The guarantee therefore rests on \emph{fully specified arithmetic}, and we treat the reproduced
cross-architecture and cross-machine comparisons of Sections~\ref{sec:experiments}--\ref{sec:distributed} as the actual
evidence --- not the mere absence of a float type.

\section{Method}
\label{sec:method}
\tritify{} is a small generic engine with one plugin crate per model family; the engine is frozen and adding a family
never touches the core. The same engine runs \emph{dense} decoder language models (Llama/Qwen/Mistral-style),
\emph{Mixture-of-Experts} decoders (OLMoE/GLM-MoE/Qwen-MoE --- one config-driven path builds dense and MoE alike, so a
new MoE model is configuration, not code), and \emph{image/video} segmentation (SAM~2.1) --- one generic exact-integer
engine across three task families.

\subsection{Representation}
A tensor is stored as integer codes $c_i \in \mathbb{Z}$ together with a single fixed-point \emph{scale}
$s = m \cdot 2^{-k}$ (an integer mantissa $m$ and a shift $k$); the real value it denotes is $c_i \cdot s$. Weights are
quantized once at build time to a balanced-ternary grid of $b$ trits ($b=12$, i.e.\ $L_W=(3^{12}-1)/2 = 265{,}720$
levels per sign, $\approx 19$ bits), $c_i = \mathrm{round}(w_i / \Delta)$ with $\Delta = \max_i|w_i| / L_W$. Activations
entering a matmul are quantized \emph{dynamically} at run time, per tensor, by their own maximum; between operations
activations are carried at the interchange level $L_A = 2^{28}$. Dynamic quantization is scale-invariant and needs no
calibration data.

\smallskip\noindent\textit{Arithmetic specification (the pinned semantics behind Claims~1 and~3).} The reference
accumulator is $i128$ (the auto-vectorized hot path accumulates in $i64$ within a proven range bound,
Section~\ref{sec:fastpaths}); a scale $m\cdot2^{-k}$ has a 64-bit mantissa $m$ (built as $\mathrm{round}(\cdot\,2^{31})$)
and a 32-bit shift $k$; requantization evaluates $\mathrm{round}(c\cdot m / 2^{k})$ as an $i128$ multiply followed by a
\emph{round-half-away-from-zero} arithmetic shift. Lookup tables use a uniform index with explicit tail handling; their
approximation error shrinks monotonically with table resolution and is reported in Section~\ref{sec:verification}. These
widths, the rounding rule, and the fixed evaluation order are what make the integer output a deterministic function of
the input across platforms --- and, in Section~\ref{sec:distributed}, across machines.

\subsection{The exact-integer data path}
\label{sec:engine-path}
\begin{itemize}
  \item \textbf{Matmul.} Accumulation is exact integer. The auto-vectorized linear kernel accumulates its
  $i32\times i32\to i64$ products in an $i64$ accumulator, which equals the exact integer sum whenever it cannot overflow
  (codes are bounded by $L_W\approx2^{18}$, so a length-$K$ sum stays below $2^{63}$ for any realistic $K$); a generic
  matmul path uses an $i128$ accumulator for cases outside that bound.
  \item \textbf{Requantization.} Scaling an integer by a fixed-point scale is $\mathrm{round}(c \cdot m \cdot 2^{-k})$
  with integer multiply and arithmetic shift; the rounding is fixed.
  \item \textbf{Residual addition.} Two tensors are brought to a common scale by integer scale composition, added
  element-wise in integer, and dynamically requantized.
  \item \textbf{Normalization (RMSNorm / LayerNorm).} The sum of squares (and mean) is an exact integer reduction; the
  per-row inverse-RMS is a scalar computed with an \emph{integer} inverse square root and folded into an integer scale.
  \item \textbf{Nonlinearities.} $\mathrm{GELU}$ and the softmax $\exp$ are integer \emph{lookup tables} built once
  offline; table resolution sets a declared, monotonically shrinking approximation error.
  \item \textbf{Attention.} Scores are an exact integer dot product scaled by the fixed $1/\sqrt{d_{\text{head}}}$; the
  softmax uses the $\exp$ table with a max-shift for stability; the rotary-position factors are precomputed into integer
  scales.
  \item \textbf{Mixture-of-Experts.} A router scores the experts in integer; the top-$k$ selection is an integer
  comparison (ties broken by index); each selected expert is an exact-integer MLP; the weighted combination is integer
  at a common scale. A shared expert and the first (dense) layers are the same integer path. The normalization $\varepsilon$ is
  resolved by one rule: a config that \emph{states} $\varepsilon$ uses that value; a config that \emph{omits} it takes
  its architecture's published \texttt{transformers} default (OLMoE-1B-7B and GLM-MoE each carry that default, with its
  source line); an architecture with \emph{no} recorded default and no stated $\varepsilon$ \emph{fails closed},
  naming the missing key. So a missing $\varepsilon$ is either resolved from the architecture's published default or
  refused --- never silently guessed. Where a recorded family's default differs from a checkpoint's own stated
  $\varepsilon$, the difference is explicit and \emph{measured}, not silent (the OLMoE case, Section~\ref{sec:depth}).
\end{itemize}

\subsection{The engine and its plugins}
Every operation implements a two-method contract: \texttt{forward\_int} (the integer run-time path) and
\texttt{forward\_ref} (a floating-point reference on dequantized inputs). A model family is a plugin that \emph{detects}
whether a checkpoint is of its family, \emph{plans} the operator graph, and \emph{builds} it from weights; a registry
selects the first plugin that recognises a checkpoint. We implement decoder-only transformers (dense and MoE) and
Segment Anything 2.1.

\subsection{Fast execution paths that preserve the output bits}
\label{sec:fastpaths}
The integer data path admits fast implementations that do not change a single output bit, because integer addition
(including two's-complement wrap-around) is associative and commutative: a reordered, multi-lane, or multi-threaded
reduction equals the scalar left-to-right one \emph{exactly} --- unlike IEEE-754, where reordering changes the low bits.
\tritify{} uses this on two axes inside a machine, and Section~\ref{sec:distributed} uses the \emph{same} property
across machines.
\begin{itemize}
  \item \textbf{SIMD.} The hot integer dot products and matmuls are written as an $i32\times i32 \to i64$ widening
  multiply--accumulate; built with \texttt{-C target-cpu=native} the compiler auto-vectorizes them to NEON
  (\texttt{smlal}) on ARM and AVX-512 on x86-64. The vectorization changes the \emph{reduction order}, not the
  accumulated value.
  \item \textbf{Multithreading.} A matmul's output rows, a convolution's output channels, and attention's heads are
  \emph{disjoint}: each output block is a function only of its index and the shared, read-only inputs. They are computed
  in parallel (a dependency-free \texttt{std::thread::scope}, keeping the engine standard-library only), writing to
  disjoint locations, so the result is order-independent and bit-identical. The worker count is a runtime knob; it
  changes only speed, never a bit.
\end{itemize}
Two facts keep these paths exact, and they must not be conflated. First, \emph{reduction order is irrelevant}: integer
addition under a fixed two's-complement wrap is associative and commutative, confirmed by a $200{,}000$-case fuzz that
deliberately includes $i64$-overflow magnitudes. Second, \emph{the sum does not actually overflow} for the dimensions
used, so each $i64$ accumulation also equals the $i128$ reference. The cross-architecture and cross-machine runs then
confirm that the fast and distributed paths reproduce the scalar reference's exact digest
(Sections~\ref{sec:perf}--\ref{sec:distributed}).

\section{Verification methodology}
\label{sec:verification}
The two-method contract is the verification mechanism. For any module or whole model, the \emph{verify gate} runs
\texttt{forward\_int} and \texttt{forward\_ref} on a fixed probe input and reports (i) the maximum absolute and relative
difference between the integer path (dequantized) and the float reference, (ii) whether the integer output is
deterministic (identical on a repeat run), and (iii) a SHA-256 of the integer output. Nothing is admitted that the gate
has not checked; an operation that cannot state its own float reference cannot ship. This turns ``is it faithful?'' from
an assertion into a test.

The gate makes Claim~2 precise: fidelity is reported \emph{to the reference}, as a tolerance. Where the operation is
exact (integer index permutations, residual addition at a common scale) the agreement is bit-exact; where it is a
quantized approximation (requantized matmuls, table nonlinearities) it is bounded. We do \emph{not} impose a pass/fail
gate on whole models: we \emph{report} measured agreement (argmax-match rate, mask IoU, maximum logit/feature error) as
diagnostics, and the measured exceptions are reported results, not gate failures.

\begin{table}[tbp]
\centering
\caption{Error of representative operations against the float reference: \textbf{M} marks an empirically measured
maximum, \textbf{D} a derived representational bound.}
\label{tab:tol}
\small
\begin{tabularx}{\textwidth}{@{}Z{1.3}Z{1.0}Z{1.1}Z{0.7}Z{0.9}@{}}
\toprule
\textbf{Operation} & \textbf{Ref / metric} & \textbf{Probe / domain} & \textbf{Gate tol} & \textbf{M / D} \\
\midrule
Fixed-point requant, full-width mantissa & f64 twin, max abs & integer scale ratios & exact & M ($0$) \\
Fixed-point requant, 8-bit mantissa & f64 twin, max abs & integer scale ratios & $\le1$ ulp & M \\
$\mathrm{GELU}$ lookup, 20-bit index & f64 $\mathrm{GELU}$, max abs & table domain & $\le10^{-4}$ & M ($4.8\!\times\!10^{-5}$) \\
Inverse-RMS (integer \texttt{isqrt}) & f64 $1/\sqrt{\cdot}$, max rel & $2\!\times\!10^{5}$ pts & $<2\!\times\!10^{-9}$ & M ($5.8\!\times\!10^{-10}$) \\
SAM~2.1 mask logits vs.\ float twin & max abs & base-plus image & diag. & M ($4.9\!\times\!10^{-2}$) \\
Qwen2.5-0.5B logits vs.\ float & max abs & 512-token probe & diag. & M ($2.1\!\times\!10^{-1}$) \\
\bottomrule
\end{tabularx}
\end{table}

\section{Experiments: the single-machine engine}
\label{sec:experiments}
We report the engine's reproducibility and fidelity on a single machine; Sections~\ref{sec:depth}--\ref{sec:distributed}
add depth and distribution. Each row names the exact model, input, hardware, comparison quantity, and a reproducing
digest (Appendix~\ref{app:digests}).

\begin{table}[tbp]
\centering
\caption{Cross-hardware reproducibility (Claim 1) and decision fidelity (Claim 2).}
\small
\begin{tabularx}{\textwidth}{@{}Z{0.7}Z{1.1}Z{1.1}Z{1.1}@{}}
\toprule
\textbf{Subject} & \textbf{Setup} & \textbf{Claim 1 (reproducibility)} & \textbf{Claim 2 (vs.\ float)} \\
\midrule
Ternary GEMM demo & $Y=XW^{\top}$, $M{=}64,N{=}768,K{=}768$; Apple M3 Max (Metal) vs.\ CPU & bit-identical CPU $=$ GPU; SHA-256 reproduced & exact integer GEMM (no float reference needed) \\
\addlinespace
Decoder LM & Qwen2.5-0.5B, seq 512; Apple M3 Max (ARM/NEON) vs.\ AMD EPYC 9454P (x86-64/AVX-512), byte-identical weights & logits SHA-256 byte-identical ARM $=$ x86-64 & greedy tokens match on tested prompts; $509/512$ argmax on the 512-token probe \\
\addlinespace
SAM 2.1 image & \texttt{hiera-base-plus}, 613-tensor checkpoint; self-check, no PyTorch & deterministic; reproducing mask SHA-256 & selected mask bit-identical ($0/65536$ px); IoU vs.\ float twin $=1.0000$ \\
\bottomrule
\end{tabularx}
\end{table}

\noindent The decoder-LM row is the strongest single-machine reproducibility evidence: two \emph{different} instruction
set architectures (ARM and x86-64) produce the \emph{same} logits digest, which is only possible because the arithmetic
is fully specified. On the SAM image the decision is exact --- both the integer path and the float twin select the same
candidate, and the selected $256\times256$ mask is pixel-identical ($0$ of $65\,536$ pixels differ) --- while the raw
logits differ by at most $4.9\times10^{-2}$.

\subsection{The fast paths are bit-identical to the reference, and measurably faster}
\label{sec:perf}
The fast paths of Section~\ref{sec:fastpaths} reproduce the scalar reference's exact digest while running substantially
faster \emph{than \tritify{}'s own scalar, single-threaded baseline}. This is \emph{not} a claim against optimized
floating-point inference; it is a demonstration that the open, verifiable reference and a fast engine are the \emph{same}
computation.

\begin{table}[tbp]
\centering
\caption{Speed of the bit-identical fast paths, relative to \tritify{}'s own scalar / single-threaded baseline. Every
cell reproduces the reference digest --- the speedups move no output bit.}
\small
\begin{tabularx}{\textwidth}{@{}Z{1.3}Z{0.85}Z{0.85}@{}}
\toprule
\textbf{Measurement} & \textbf{Speedup} & \textbf{Bit-identity} \\
\midrule
SIMD vs.\ vectorizers-off, hot kernel ($896{\times}896$, $4096$ rows) & $4.76\times$ (M3/NEON); $4.68\times$ (EPYC/AVX-512) & same digest; cross-ISA checksum equal \\
\addlinespace
SIMD\,$\times$\,threads: Qwen2.5-0.5B \texttt{forward\_int}, seq~512 & $11.8\times$ ($133.7\,\mathrm{s}\!\to\!11.4\,\mathrm{s}$) & all cells: same logits hash \\
\addlinespace
SIMD\,$\times$\,threads: SAM~2.1 \texttt{base-plus}, end-to-end wall & $7.0\times$ ($237.7\,\mathrm{s}\!\to\!34.0\,\mathrm{s}$) & all cells: same mask-logits hash \\
\bottomrule
\end{tabularx}
\end{table}

\noindent The point is the last column: the \emph{same reproducing hash} appears in every cell --- scalar and
vectorized, one thread and sixteen --- so a deployment may choose the ISA and the thread count purely for performance,
with the scalar reference as an oracle that re-derives the exact bytes. Section~\ref{sec:distributed} extends exactly
this ``choose freely, the answer never moves'' property from the cores of one machine to a network of machines.

\subsection{Footprint}
\label{sec:footprint}
Transforming Qwen2.5-0.5B ($494$\,M parameters) emits a self-contained KPI card: the original is $942$\,MiB (BF16); the
transformed model is $\approx 1.15$\,GiB ($b{=}12$, $\approx 20$ bits/weight packed) $={}1.25\times$ the fp16 size, with
a peak conversion working set of $5.80$\,GiB and a convert-plus-self-verify time of $13.6$\,s. This is deliberately
\emph{not} a compression result: \tritify{} trades size, memory, and per-device speed for exactness, determinism, and
verifiability, and is $\sim$fp16-class on footprint.

\section{Fidelity at depth}
\label{sec:depth}
Section~\ref{sec:experiments}'s fidelity evidence is end-to-end (final logits / mask) or on short slices. Because the
decision-match of Claim~2 is an empirical property, we characterise how the integer path tracks the float reference
\emph{as layers accumulate}, on a fixed $32$-token probe, measuring after every layer the residual stream's per-position
cosine, each MoE layer's top-$k$ routing agreement, and, at the end, the final-normed hidden state and logits. We run
two models: OLMoE-1B-7B (all sixteen layers --- the calibration, a build the anchors call known-correct at full depth)
and the first sixteen layers of GLM-4.5-Air (a $106$-billion-parameter frontier MoE; this is the deepest slice a single
node holds in fp32 for the reference). The float reference is \texttt{transformers} at a pinned revision, in fp32 on CPU.

\begin{table}[tbp]
\centering
\caption{Per-layer fidelity of the integer path against the fp32 reference, measured after every layer; no fidelity
cliff. Both sides run at each model's own \texttt{transformers}-configured $\varepsilon$ --- OLMoE-1B-7B at
$1\mathrm{e}{-}5$ (the engine's family default was aligned to this shipped value; see text) and GLM-4.5-Air at its
stated $1\mathrm{e}{-}5$ --- so each column is decision fidelity to that \emph{shipped} checkpoint, unmodified.}
\label{tab:depth}
\small
\begin{tabularx}{\textwidth}{@{}Z{1.2}Z{0.9}Z{0.9}@{}}
\toprule
\textbf{Measure (over 16 layers)} & \textbf{OLMoE-1B-7B} & \textbf{GLM-4.5-Air (first 16)} \\
\midrule
Residual cosine, median & $0.9999991$ & $0.9996717$ \\
Residual cosine, min (final layer) & $0.99617$ & $0.9873843$ \\
Residual cosine, worst position (any layer) & $0.99617$ & $0.9735$ (recovers to $0.9988$) \\
Final logits cosine, median & $0.9999998$ & $0.9995361$ \\
Router top-$k$ exact agreement & $508/512$ & $448/480$ \\
Largest layer-to-layer median cosine drop & $3.2\times10^{-7}$ & $<10^{-3}$ \\
No-cliff gate (per-layer drop $<10^{-3}$; routing $\ge 24/32$) & holds & holds \\
\bottomrule
\end{tabularx}
\end{table}

\begin{figure}[tbp]
\centering
\includegraphics[width=0.93\linewidth]{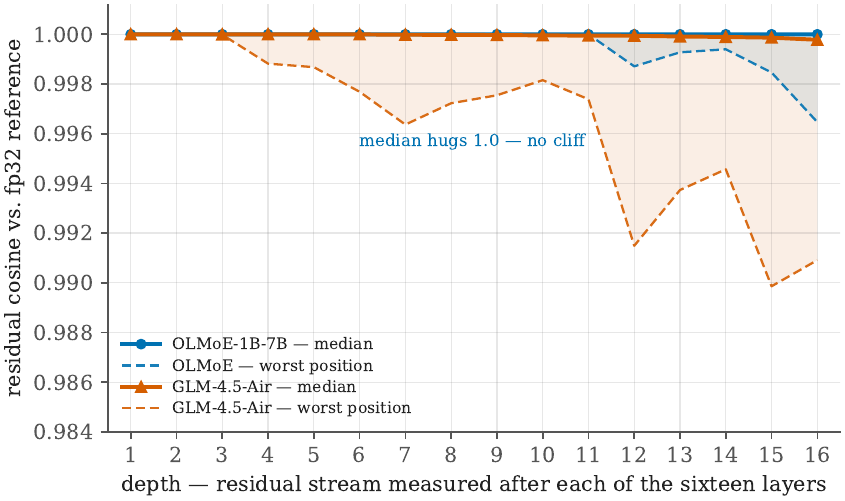}
\caption{Fidelity does not cliff with depth. Per-layer residual cosine to the fp32 reference, both sides at the
checkpoint's own $\varepsilon$ (zoomed axis). Solid: the per-position \emph{median}, near $1.0$ throughout.
Dashed/shaded: the single \emph{worst} probe position, dipping to $\approx 0.996$ (OLMoE) and $\approx 0.99$ (GLM) at
the deepest layers and no further --- a bounded spread, not a collapse.}
\label{fig:nocliff}
\end{figure}

\noindent The integer arithmetic stays float-faithful through depth on both models --- a calibration (OLMoE) and a
frontier MoE (GLM-4.5-Air) --- with no layer at which fidelity collapses (Figure~\ref{fig:nocliff}). We report, not hide, the facts that complicate
the picture. Of the $480$ GLM routing decisions ($15$ MoE layers $\times$ $32$ probe positions), $32$ had a top-$k$ set
that differed from the reference --- near-ties, not a systematic divergence; one probe position in particular flipped an
expert at several layers, dipping its residual cosine to $0.9735$ at one layer and recovering to $0.9988$ by layer~16.
GLM's end-of-slice $1-\text{median cosine}$ is about $3\times10^{-4}$ --- below a $0.9999$ figure we had \emph{expected}
but above every gate. Both columns are decision fidelity to the \emph{shipped} checkpoint. \tritify{} builds the OLMoE
family at \texttt{transformers}' configured $\varepsilon=1\mathrm{e}{-}5$, so the integer and float sides share that
checkpoint's own $\varepsilon$ and agree on $508/512$ routing decisions. We report how we got here, rather than hide it:
an earlier build defaulted the family to $1\mathrm{e}{-}6$ when the checkpoint's config omits the key, and against the
shipped $\varepsilon$ that build agreed on only $0.455$ of routing decisions --- an $\varepsilon$-\emph{configuration}
difference, not integer-arithmetic error. We aligned the engine's family default to each architecture's
\texttt{transformers} default (by the single rule of Section~\ref{sec:engine-path}: a \emph{missing} $\varepsilon$ is
taken from a \emph{recorded} family's \texttt{transformers} default, and an unrecorded family with no stated value fails
closed --- never silently guessed), re-anchored the OLMoE hashes accordingly, and the figures above are measured on that
corrected build. The GLM slice
never needed this --- its config states $\varepsilon$ explicitly. \textbf{Honest scope:} this is sixteen of GLM-4.5-Air's
forty-six layers --- the depth a single node holds in fp32; the curve past layer sixteen is an extrapolation, and the
full-depth property we establish in Section~\ref{sec:distributed} is \emph{bit-identity}, which is exact, not fidelity
to float, which is measured.

\section{Deterministic routing introspection}
\label{sec:routing}
Because the forward is exact integer, a Mixture-of-Experts routing decision --- which top-$k$ experts each token
selects at each layer --- is a deterministic, byte-reproducible function of (weights, input). A read-only trace hook
records it \emph{without changing a bit of the forward}: the verify gate stays green and the output hash is unchanged,
so the same model and input yield the same routing trace on every machine. At a fixed numeric configuration and
execution schedule, this trace is \emph{identical across every backend we tested} --- the same conditions under which
Claim~1 holds. A float engine gives no such guarantee by default: near a routing tie its expert selection can differ
run-to-run or across hosts (the same non-associativity that makes float \emph{output} non-reproducible across
environments), so a float routing trace is not portable without the reduction-pinning effort that makes float output
reproducible. The same generic path serves any MoE family with no model-specific code; we exercise it on OLMoE-1B-7B
($64$ experts) and Qwen3-30B-A3B ($128$ experts).

\begin{figure}[tbp]
\centering
\includegraphics[width=0.95\linewidth]{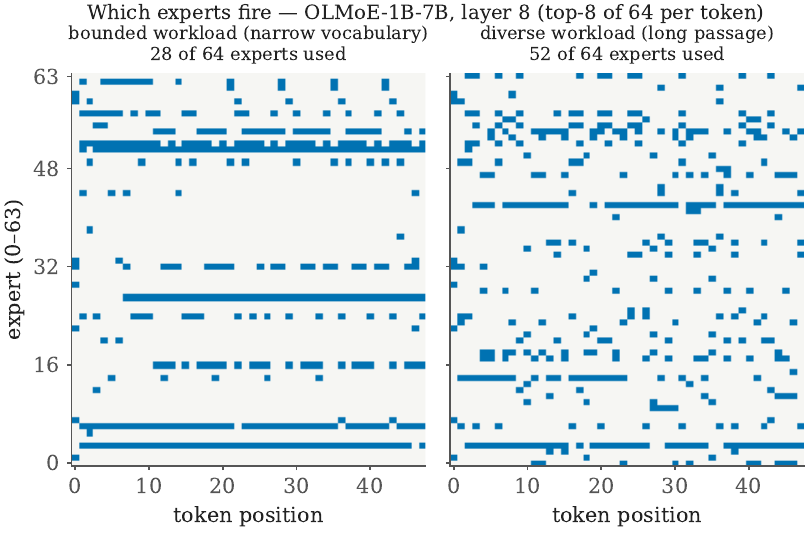}
\caption{Which experts fire, read directly off the exact-integer forward (OLMoE-1B-7B, one MoE layer; each column a
token, each row one of the $64$ experts, a cell marked when that expert is in the token's top-$8$). A bounded,
narrow-vocabulary workload reuses a small set of experts --- the horizontal bands, $28$ of $64$ here --- while a
diverse passage spreads across most of them ($52$ of $64$). Because the routing decision is deterministic, this trace
is exactly reproducible. Generated from a committed route trace.}
\label{fig:heatmap}
\end{figure}

\begin{figure}[tbp]
\centering
\includegraphics[width=0.9\linewidth]{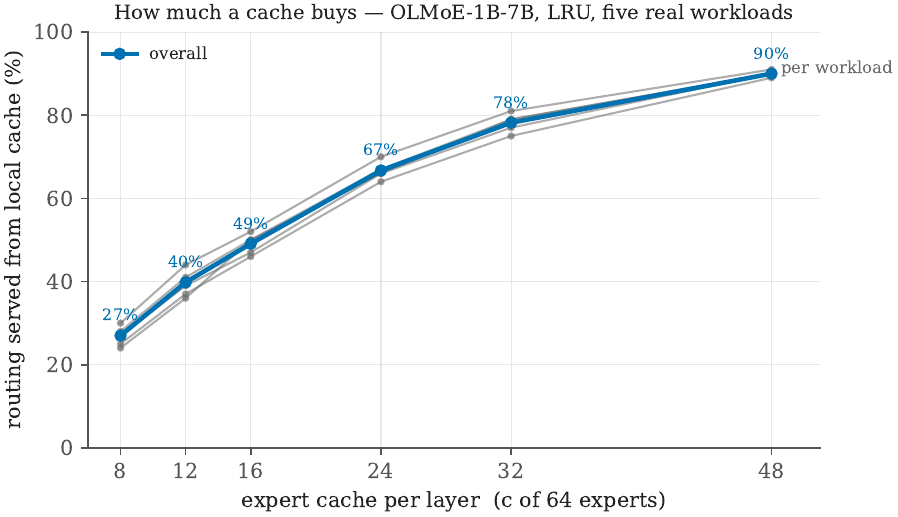}
\caption{How much a local expert cache buys, as a relationship rather than a single number: the fraction of routing an
LRU cache serves locally against the cache size, for five real OLMoE-1B-7B workloads and their mean. This is
out-of-sample --- the cache is \emph{not} fitted to the probe it serves. A modest cache ($c=24$ of $64$ experts per
layer) already serves about two thirds; the curve is workload-dependent, bounded by each workload's own vocabulary.
Records released with the artifact.}
\label{fig:cache}
\end{figure}

What the trace shows (records released with the artifact) is that the routing a model produces \emph{tracks the
input's vocabulary}: a single repeated token selects a handful of experts with near-perfect temporal locality, while
maximally diverse text spreads across almost all of them (Figure~\ref{fig:heatmap}), and a bounded real-world workload
(one user or domain) sits in between --- with a working set small enough that an edge cache of experts pays off (Figure~\ref{fig:cache}). On the workloads we examined
this working set grew with the input's vocabulary rather than with the model's expert count; \emph{if} that pattern
holds more broadly it would make larger MoEs more cache- and placement-friendly for bounded workloads, not less. We
offer this as an observation on the cases we ran, not a law --- context, token order, and architecture also shape the
hidden states that drive routing, and we have not isolated them. Section~\ref{sec:distributed} shows one concrete use,
with its cache size and in-sample scope stated there. The claim this paper makes is narrow and exact: the introspection
is \emph{deterministic} --- the same routing, reproducibly, on any machine at a fixed configuration --- which is what
makes it trustworthy as a measurement rather than an impression.

\section{Verifiable distributed inference}
\label{sec:distributed}
\begin{figure}[tbp]
\centering
\includegraphics[width=0.93\linewidth]{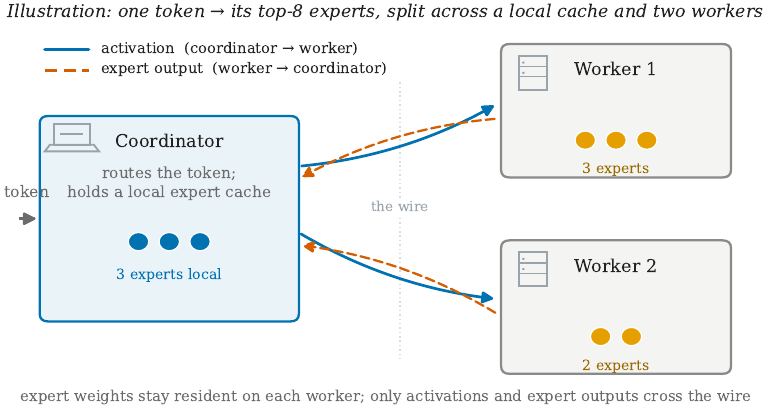}
\caption{The distributed seam (schematic). The coordinator routes each token locally and sends its \emph{activation}
to the worker holding each selected expert, which returns the expert's \emph{output vector}; expert \emph{weights} stay
resident and never cross the wire (so per-token traffic is kilobytes). The split shown ($3$ local, $3{+}2$ remote) is
an example --- the real placement is the oracle's. Recombination is placement-independent, so the output carries one
SHA-256, identical on every node (Claim~3).}
\label{fig:seam}
\end{figure}

Claim~3 says the output does not depend on \emph{where} each operation runs. This section is its demonstration and its
consequence. The enabler is structural and we state it plainly: because every machine computes the same exact integers,
\emph{which} machine runs which layers --- or which experts, or which rows of a matmul --- has \emph{zero effect on the
output bits}, verified by a single SHA-256 (Figure~\ref{fig:seam}). Standard float inference does not do this by default: CPU-float, GPU-float,
and a second CPU's float differ bit-for-bit, so a split float result depends on the split and is not reproducible
without special effort (a reduction-pinned BLAS fixes summation order but not a whole heterogeneous model).

\subsection{A whole model across six machines, bit-identically}
\label{sec:fleet6}
\begin{figure}[tbp]
\centering
\includegraphics[width=0.52\linewidth]{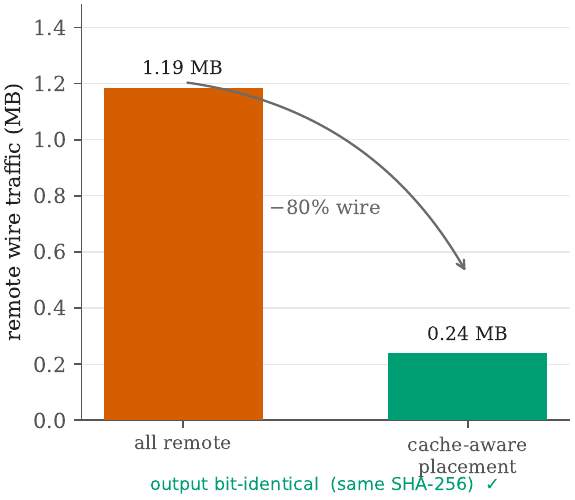}
\caption{What cache-aware placement saves, at no cost to correctness: remote wire traffic for the loopback probe above,
all-remote versus the cache-aware placement, with the output bit-identical (the same SHA-256) in both. In-sample cache
as stated in the text. Records released with the artifact.}
\label{fig:placement}
\end{figure}

We split OLMoE-1B-7B (all sixteen layers; $64$ expert networks per MoE layer) across \emph{six} live machines of two
instruction sets and three locations: a MacBook~Pro (arm64) as the \emph{coordinator} that routes each token locally,
and five \emph{workers} holding the experts between them --- a 96-core cloud server (x86-64), two fanless x86 mini-PCs,
and two Raspberry~Pis (aarch64) --- reached over QUIC. Each machine holds only its shard of experts; for each token the
coordinator routes locally, then sends the token's hidden-state \emph{activation} to the worker holding each selected
expert, which evaluates those experts and returns their \emph{output vectors}; the coordinator combines them. The
expert \emph{weights} stay resident on their worker and never cross the wire --- only activations (out) and the experts'
output vectors (in) do, which is why the per-token traffic is kilobytes, not the gigabytes an expert's weights would be.
The recombination is
placement-independent, so the distributed gather reproduces the single-machine output exactly: the whole model carries
the single-machine hash \texttt{3dadb4f1\ldots} both ways. (This fleet run was recorded on the build that preceded the
OLMoE $\varepsilon$ alignment of Section~\ref{sec:depth}, so its hash is that model's $1\mathrm{e}{-}6$ build;
bit-identity is a property of the shared integer path and is independent of the $\varepsilon$ value.) The $8$-token
probe cost $498$ expert round-trips, $4.10$\,MB out / $8.43$\,MB in, $35.8$\,s of wall time, with every expert fetched
remotely. On a separate two-layer loopback of the same probe we measured what a cache buys: a cache-aware placement ---
a resident set of $24$ of the $64$ experts per layer, chosen \emph{in-sample} as that probe's own most-used experts ---
served $85\%$ of its routing ($109/128$ routed experts) locally and cut remote wire traffic by about $80\%$, at
identical output (Figure~\ref{fig:placement}). That measures the mechanism's locality on this one workload, not an
out-of-sample generalization, which we do not claim here. ``Different silicon, different locations, same bits.''

\subsection{A frontier model no single node can hold}
\label{sec:flagship}
\begin{figure}[tbp]
\centering
\includegraphics[width=0.9\linewidth]{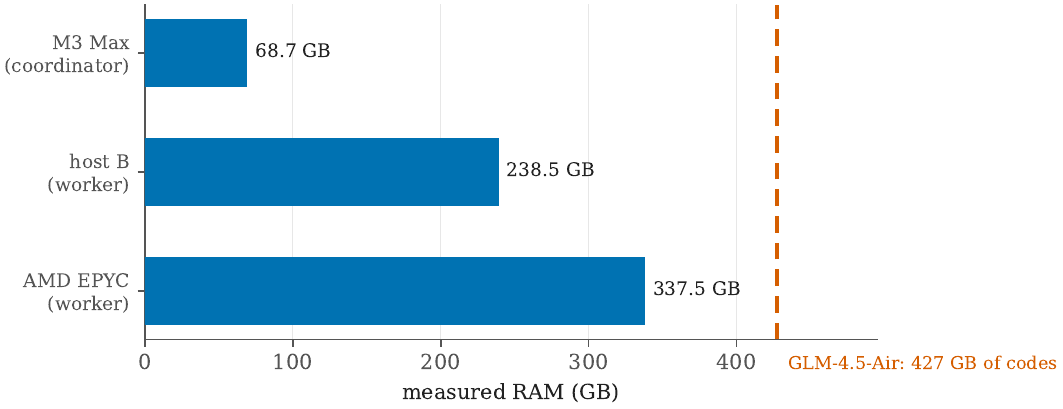}
\caption{No single node holds the model. Measured RAM of each machine versus GLM-4.5-Air's $427$ GB of codes (dashed).
Every node is left of the line, yet the two workers together ($576$ GB) exceed the $399$ GB of routed codes they hold
--- so the fleet runs a model no single node can hold in RAM at once (the single-host reference runs the \emph{same}
model by streaming). Labels anonymized as in the text.}
\label{fig:capacity}
\end{figure}

We then ran the capacity case: GLM-4.5-Air, $106{,}852{,}251{,}264$ parameters, $427{,}409{,}005{,}056$ bytes of integer
codes --- more than the memory of \emph{any} machine in the run (Figure~\ref{fig:capacity}). The fleet was three machines: an Apple~M3~Max
coordinator ($68.7$\,GB) holding everything \emph{but} the routed experts ($28{,}782{,}352{,}896$ bytes of codes, and it
declares \texttt{holds\_full\_model: false}), and two workers holding the $128$ routed experts of each MoE layer between
them --- an AMD~EPYC (experts $0$--$74$, $233{,}570{,}304{,}000$ bytes) and a second host (experts $75$--$127$,
$165{,}056{,}348{,}160$ bytes). The workers build their experts' codes on demand from the bf16 checkpoint under a memory
budget; a cached, evicted, or rebuilt expert produces the \emph{same} codes, because quantization is a pure function of a
tensor's bytes.

Asked ``What is the capital of France? Answer in one sentence.'' in GLM-4.5's chat template, the fleet answered
\texttt{"The capital of France is Paris."} --- and it is a single host's answer to the bit. A single-host reference on
one machine, streaming each expert from the checkpoint as it is needed, produced the identical result: the same seven
tokens, the same generated-logits hash \texttt{564f495c\ldots}, and the same hidden-state hash at \emph{every one of the
forty-six layers}.

\begin{table}[tbp]
\centering
\caption{The flagship run (GLM-4.5-Air, 46 layers, $427$\,GB of codes) across a fleet none of which can hold its full
codes in RAM, against a single-host streamed reference. Every memory figure is measured on its node; ``resident'' is
measured RAM, distinct from the assigned expert ranges.}
\label{tab:flagship}
\small
\begin{tabularx}{\textwidth}{@{}Z{0.9}Z{0.8}Z{1.3}@{}}
\toprule
\textbf{Property} & \textbf{Value} & \textbf{Meaning} \\
\midrule
Model codes & $427{,}409{,}005{,}056$ B & exceeds every node's memory in this run \\
Coordinator held & $28{,}782{,}352{,}896$ B & router-only; \texttt{holds\_full\_model: false} \\
Workers resident (sum) & $398{,}626{,}652{,}160$ B & codes \emph{resident} after warm-up (measured) $=$ the assigned routed experts' codes; on-demand build, whole share fit here \\
Node memory (each) & M3 $68.7$ · host~B $238.5$ · EPYC $337.5$ GB & each $<427.4$\,GB: no node holds the full codes in RAM \\
Bit-identity & \texttt{564f495c\ldots} both ways & distributed $=$ single-host streamed reference, all 46 layers \\
Cost & $2{,}504$ round-trips, $633.6$\,s total & $\sim$90 round-trips per decode \emph{position} ($45$ MoE $\times$ $2$); $2{,}504$ over $21$ prompt positions $+$ $7$ tokens; total wall, not a decode rate \\
\bottomrule
\end{tabularx}
\end{table}

\begin{calloutblue}
\noindent This is the demonstration of Claim~3 at frontier scale: a model whose full integer codes ($427$\,GB) exceed
every machine's RAM in the run generates a correct, single-host-identical answer across a fleet, provably, by one hash.
The win is running such a model with each node's resident working set bounded --- \emph{not} an exclusive ``runs at
all'': the single-host reference runs the \emph{same} model by streaming experts from disk, so the cluster's practical
advantage over single-host streaming (resident set, latency, disk~I/O) is a separate, measurable quantity we do not
quantify here. Distribution adds round-trips --- about $90$ per decode \emph{position} ($45$ MoE layers $\times$ $2$
workers), $2{,}504$ in total over the prompt's $21$ positions plus the $7$ generated tokens, in $633.6$\,s of total wall
time (setup, prefill and decode not separated here, so no decode-token rate is implied) --- and we do not present it as
faster than a single host that could hold the model.
\end{calloutblue}

\subsection{What bit-identity unlocks: verifiable offload}
\label{sec:offload}
The same property makes an \emph{untrusted} server \emph{checkable} --- a direction we frame carefully, because what we
have demonstrated is the agreement of \emph{honest} distributed execution with a reference, not a verification protocol
against a dishonest one. The server (or fleet) is the fast engine; the open scalar reference --- local, or any third
party --- is the referee. Full local recomputation gives an exact comparison; recomputing a random sample of operations
or tokens gives only a detection \emph{probability}, set by the sampling rule. Turning that into a security guarantee
needs a specified selection, a binding of the intermediate states, and an explicit attacker model --- \emph{none of
which we provide here}. A fully specified, analysed verification protocol would be its own contribution; Slalom~\cite{slalom}
is relevant prior work on verifiable neural-network inference. With that caveat stated, three capabilities the property
\emph{enables} (each purely because the engine is deterministic):
\begin{itemize}
  \item \textbf{Verify / audit (a direction).} Recompute $N$ random ops or tokens locally and check the hash. This
  \emph{detects} a dishonest server with a probability set by $N$ and the (still-to-be-specified) sampling rule; it is a
  statistical or after-the-fact check, not a proof, and depends on the attacker model and state-binding noted above.
  \item \textbf{Same-state handoff.} For the \emph{same} model under the \emph{same} numeric configuration and
  batching/execution schedule, a KV-cache computed on one node is bit-identical to the one another node would produce,
  so a session can move between hosts (e.g.\ local to remote for longer context) with a provably identical continuation
  --- \emph{once} the incremental KV-cache decode path exists, which today is future work
  (Section~\ref{sec:limitations}). The equality is within one model \emph{and one schedule}: a cache is \emph{not}
  transferable to a different or larger model, nor across a different batching/execution schedule (the dynamic-quantization
  schedule dependence of Section~\ref{sec:limitations}).
  \item \textbf{Deterministic cache key.} Output is a pure function of (weights, inputs, and the fixed batching/execution
  schedule) with an \emph{exact} key, so a prefix/KV cache populated on one backend is numerically identical on another
  \emph{at the same schedule}. Float systems already cache prefixes keyed by tokens (e.g.\ vLLM~\cite{vllm}); what
  bit-identity adds is cross-backend \emph{value} equality, so a cache is reusable across mismatched hardware --- not
  that float caching is impossible.
\end{itemize}
This is a third option between all-local (private but memory-bound) and all-cloud (powerful but trust-me): cloud or
fleet capacity with a reproducible, independently checkable result.

\begin{calloutwarn}
\noindent\textbf{Honest limits (stated plainly, not sold as hype).} \emph{Verifiability is not confidentiality.} A server that
runs the layers \emph{sees} the activations; splitting early layers local and later layers remote \emph{improves}
privacy but does not guarantee it, and row-splitting a layer across nodes hides nothing (each node needs the full layer
input). A genuine confidentiality claim needs its own mechanism (a deep cut whose inversion is \emph{measured} to be
hard, added noise, or MPC/HE) --- it does not follow from bit-identity. \emph{Verification costs recomputation:} full
re-execution erases the offload benefit, so the value is the sampling/audit model, a statistical or offline guarantee,
not free total verification. \emph{Bandwidth is the real cost:} activations (or experts) cross the wire every token, so
a slow link or a weak device can make the distributed run slower than not distributing at all --- prefer wired,
capacity-bound settings. \emph{Determinism simplifies correctness; it does not supply the protocol.} Because a slice
is bit-identical wherever it runs, re-executing it after a fault reproduces the exact bits --- but consistent cluster
state, fencing a slow or partitioned node, and resuming a session still require a correct distributed protocol;
determinism makes that protocol's \emph{results} reproducible, not the protocol unnecessary.
\end{calloutwarn}

\section{Limitations}
\label{sec:limitations}
\begin{itemize}
  \item \textbf{No speed claim against float.} The integer path is not faster than optimized float inference on general
  CPUs/GPUs; the value is determinism, portability, and verifiability. The intra-machine speedups
  (Section~\ref{sec:perf}) are relative to \tritify{}'s \emph{own} scalar baseline, and distribution
  (Section~\ref{sec:distributed}) \emph{adds} latency --- its win is capacity and verifiability.
  \item \textbf{Tolerance, not bug.} Claim~2 is agreement \emph{within a declared tolerance}; the raw logits are not
  bit-equal to the reference. On the trials we ran the decisions matched, with the measured exceptions reported.
  \item \textbf{Bit-identity vs.\ fidelity, at depth.} Across the fleet (Section~\ref{sec:distributed}) the distributed
  run is \emph{bit-identical} to the single-host reference at full depth (all 46 layers) --- an \emph{exact} property.
  Fidelity \emph{to a float model} is a separate, \emph{measured} property, and we measure it to sixteen layers
  (Section~\ref{sec:depth}); we do not compare the full 46-layer answer to a float model, because no node in the run
  holds GLM-4.5-Air in fp32.
  \item \textbf{Scope of determinism (batch / schedule).} Claim~1 is reproducibility for a \emph{fixed input and a fixed
  batching / execution schedule}. Because activation quantization is dynamic per tensor, a different batch composition or
  prefill/decode split can change a tensor's maximum and hence its quantization; we \emph{measured} this on
  Qwen2.5-0.5B (full prefill vs.\ incremental decode) and the integer path is \emph{not} schedule-invariant, though every
  argmax decision was preserved on that probe and the float reference is exactly invariant --- locating the effect in the
  dynamic per-tensor quantization. A per-token / static scaling choice attains invariance (as Barry's \texttt{detllm}
  does); adopting it, together with the KV-cache incremental-decode path, is future work.
  \item \textbf{Confidentiality is not claimed.} See Section~\ref{sec:offload}: bit-identity gives verifiability, not
  privacy.
  \item \textbf{Scale validated, not asymptotic.} We validate dense 0.5B--1.5B LMs, OLMoE-1B-7B, SAM~2.1
  \texttt{base-plus}, and GLM-4.5-Air across a fleet; an exhaustive platform matrix --- including a
  compiler/toolchain-version matrix run as its own reproducibility experiment --- and larger single-node fp32 fidelity
  references are future work. The GPU backend's bit-identity to the CPU path is shown for the ternary GEMM on Apple~M3
  Metal, not yet for whole models.
  \item \textbf{Artifact completeness.} This is a preprint draft: the public repository URL, several checkpoint and input
  hashes, build configurations, exact run commands, and full run logs are to be fixed at release. An output hash enables an identity
  comparison but does not by itself certify a correct model implementation.
\end{itemize}

\section{Related work}
\label{sec:related}
Fully integer transformer inference --- including the nonlinear operations --- has direct precedent: I-BERT~\cite{ibert},
I-ViT~\cite{ivit}, and I-LLM~\cite{illm} quantize attention, softmax, GELU, and normalization to integer or dyadic
approximations. Post-training quantization and low-bit / ternary weights (the BitNet line~\cite{bitnet}), incoherence
rotations (QuIP\#~\cite{quipsharp}, SpinQuant~\cite{spinquant}), and integer kernels (\texttt{gemmlowp} / TFLite
integer-arithmetic-only inference~\cite{jacob2018}) are the arithmetic lineage; the determinism concern connects to
reproducible computing (reproducible BLAS~\cite{reproblas}) and the non-associativity of IEEE-754. The closest public
project, \texttt{int-llm}~\cite{intllm}, is a pure-integer C stack reporting byte-reproducible output across platforms;
\tritify{} matches that pure-integer run-time core and adds breadth (a decoder-LM family \emph{and} SAM~2.1), a uniform
run-time \texttt{forward\_int}/\texttt{forward\_ref} contract, an Apache-2.0 oracle, and a single portable GPU shader.
Barry's \texttt{detllm}~\cite{detllm_barry} runs the entire forward pass of Qwen3-0.6B in exact integer arithmetic and
obtains one int8 logits hash identical across a broader hardware matrix than ours (two NVIDIA GPU architectures and
AMD/Intel/Apple CPUs), and crucially \emph{attains} the batch- and schedule-order invariance we currently measure and
lack; it is, however, a single small decoder with no vision, and the repository carries no license at the time of
writing. A distinct line of \emph{float}-reproducibility checkers (one also named ``detllm''~\cite{detllm_cerruti})
verifies ordinary floating-point inference rather than replacing it.

Distributed inference across commodity or volunteer hardware --- pipeline and tensor parallelism, and systems such as
Petals~\cite{petals} that serve large models over a network --- is well established for \emph{throughput} and
\emph{capacity}. What is new here is not distribution but its \emph{provability}: because the engine is bit-identical
regardless of where an operation runs, a distributed gather equals the single machine \emph{exactly} and is checkable by
one hash, and an offloaded computation can be \emph{verified} rather than trusted. Float distributed-inference systems
deliver capacity and speed; what they do not provide \emph{by default} is a reproducible, independently checkable
result (reproducible float is achievable at a cost, e.g.\ a reduction-pinned BLAS~\cite{reproblas}, but is not the norm
in these stacks). That default, built-in verifiability --- not the distribution itself --- is our contribution.

\needspace{10\baselineskip}%
\section{A separate commercial product}
\tritify{} --- the engine, the oracle, and the per-operation verification --- is open source. Pelloran Inc.\ separately
offers the \emph{system} built on it: an orchestrator for distributed and heterogeneous deployment (placement,
load-balancing, comms, fault tolerance, and the audit protocol) and verifiable-inference / compliance services, together
with the independent \emph{Generation~5 Embedding} retrieval product. This mention is for disclosure; the engine and the
reproducibility, fidelity, and bit-identity results in this paper stand on their own and depend on none of it.

\section{Reproduction}
On release, from the public repository --- with no account and no network --- the determinism result and the
per-operation gate reproduce directly:
\begin{lstlisting}[language=bash]
git clone https://github.com/pelloran-inc/tritify
cd tritify
cargo test --workspace      # every op: integer path checked against its float reference
cargo run --release         # zero-float ternary GEMM; GPU == CPU; prints a reproducing SHA-256
\end{lstlisting}
Running the last command again, or on another machine, yields the identical digest. The model-level and distributed
experiments additionally require the corresponding public checkpoints and the fleet configuration; each run records its
nodes' measured memory, the per-layer and generated hashes, and the exact command, so that on release a third party can
reproduce the bytes and re-check the single-host identity.

\section*{Availability}
\tritify{} is licensed under Apache-2.0. The repository --- the artifact of this paper --- contains the engine, the
model-family plugins, the self-verification gate, the determinism demonstration, the fidelity-at-depth harness, and the
distributed coordinator/worker with the exact commands that regenerate the results above. \emph{This is a preprint
draft: several checkpoint and input hashes, the build configurations, and the full run logs
that complete the reproducibility artifact are fixed at release and are not all present in this draft.}

\appendix
\section{Reproducing digests}
\label{app:digests}
Each result carries a reproducing integer-output SHA-256 (truncated here; full-length digests are fixed at release).

\begin{center}
\small
\begin{tabularx}{\textwidth}{@{}Z{1.2}Z{0.8}@{}}
\toprule
\textbf{Artifact (host, backends)} & \textbf{Reproducing SHA-256} \\
\midrule
Single-machine determinism demonstration (Apple~M3~Max) & \texttt{de67ba75\ldots} \\
SAM~2.1 image self-check (no PyTorch) & \texttt{7177a5f5\ldots} \\
Qwen2.5-0.5B decoder-LM, end-to-end --- Apple~M3~Max (ARM/NEON) \emph{and} AMD~EPYC~9454P (x86-64/AVX-512), on
byte-identical weights, scalar and vectorized, one and all cores & \texttt{6c139feb\ldots} \\
OLMoE-1B-7B --- six-machine distributed run $=$ single machine & \texttt{3dadb4f1\ldots} \\
\quad loopback anchor for the above & \texttt{6ae7e544\ldots} \\
OLMoE-1B-7B --- corrected-$\varepsilon$ build, bit-identical across ARM and x86 & \texttt{74005166\ldots}\,/\,\texttt{505cb108\ldots} \\
GLM-4.5-Air flagship --- distributed $=$ single-host reference, every one of the 46 layers & \texttt{564f495c\ldots} \\
\bottomrule
\end{tabularx}
\end{center}

\noindent The six-machine OLMoE run predates the OLMoE $\varepsilon$ alignment of Section~\ref{sec:depth}, so its two
anchors (\texttt{3dadb4f1\ldots} and \texttt{6ae7e544\ldots}) are that model's $1\mathrm{e}{-}6$ build; the
bit-identity demonstrated does not depend on the $\varepsilon$ value. The full-length
digests are fixed at release.


\begin{thebibliography}{99}
\bibitem{ibert} S.~Kim, A.~Gholami, Z.~Yao, M.~W.~Mahoney, and K.~Keutzer. I-BERT: Integer-only BERT quantization. In \emph{Proc.\ ICML}, PMLR vol.~139, pp.~5506--5518, 2021. arXiv:2101.01321.
\bibitem{ivit} Z.~Li and Q.~Gu. I-ViT: Integer-only quantization for efficient vision transformer inference. In \emph{Proc.\ ICCV}, pp.~17065--17075, 2023. arXiv:2207.01405.
\bibitem{illm} X.~Hu et al. I-LLM: Efficient integer-only inference for fully-quantized low-bit large language models. arXiv:2405.17849, 2024.
\bibitem{jacob2018} B.~Jacob et al. Quantization and training of neural networks for efficient integer-arithmetic-only inference. In \emph{Proc.\ CVPR}, pp.~2704--2713, 2018. arXiv:1712.05877.
\bibitem{bitnet} S.~Ma et al. The era of 1-bit LLMs: All large language models are in 1.58 bits. arXiv:2402.17764, 2024.
\bibitem{quipsharp} A.~Tseng, J.~Chee, Q.~Sun, V.~Kuleshov, and C.~De~Sa. QuIP\#: Even better LLM quantization with Hadamard incoherence and lattice codebooks. In \emph{Proc.\ ICML}, PMLR vol.~235, 2024. arXiv:2402.04396.
\bibitem{spinquant} Z.~Liu et al. SpinQuant: LLM quantization with learned rotations. In \emph{Proc.\ ICLR}, 2025. arXiv:2405.16406.
\bibitem{reproblas} J.~Demmel and H.~D.~Nguyen. Fast reproducible floating-point summation. In \emph{Proc.\ 21st IEEE Symposium on Computer Arithmetic (ARITH)}, 2013.
\bibitem{intllm} N.~Micic. int-llm: Integer-only LLM inference in pure C. GitHub repository, 2026.
\bibitem{detllm_barry} N.~Barry. DetLLM: Fully deterministic LLM inference. GitHub repository, 2026 (read 2026-10-29).
\bibitem{detllm_cerruti} T.~Cerruti. detllm: Deterministic-mode checks for LLM inference. GitHub repository.
\bibitem{petals} A.~Borzunov et al. Petals: Collaborative inference and fine-tuning of large models. In \emph{Proc.\ ACL (demo)}, 2023. arXiv:2209.01188.
\bibitem{slalom} F.~Tram\`er and D.~Boneh. Slalom: Fast, verifiable and private execution of neural networks in trusted hardware. In \emph{Proc.\ ICLR}, 2019. arXiv:1806.03287.
\bibitem{vllm} W.~Kwon, Z.~Li, S.~Zhuang, et al. Efficient memory management for large language model serving with PagedAttention. In \emph{Proc.\ SOSP}, 2023. arXiv:2309.06180.
\end{thebibliography}
\end{document}